\documentclass[10pt,twocolumn,letterpaper]{article}

\usepackage[pagenumbers]{cvpr}

\newcommand{\cmark}{\ding{51}} 
\newcommand{\xmark}{}

\usepackage[T1]{tipa}

\definecolor{cvprblue}{rgb}{0.21,0.49,0.74}

\PassOptionsToPackage{
  pagebackref,
  breaklinks,
  colorlinks,
  allcolors=cvprblue
}{hyperref}

\usepackage{skeldoc}

\skelset{hypersetup=colorlinks}

\usepackage{tabularx}
\usepackage{pifont}
\usepackage{multirow}
\usepackage{graphicx}
\usepackage {zebra-goodies}
\usepackage{bm}
\usepackage{subcaption}

\def\paperID{}
\def\confName{}
\def\confYear{}

\title{
M-PhyGs: Multi-Material Object Dynamics from Video
}

\author{Norika Wada \qquad Kohei Yamashita \qquad Ryo Kawahara \qquad Ko Nishino\\
Graduate School of Informatics, Kyoto University,
Kyoto, Japan\\
{\tt\small \url{https://vision.ist.i.kyoto-u.ac.jp/research/m-phygs/}}
}

\begin{document}

\twocolumn[{
  \maketitle
  \vspace{-4ex}
  \begin{center}
    \includegraphics[width=\linewidth]{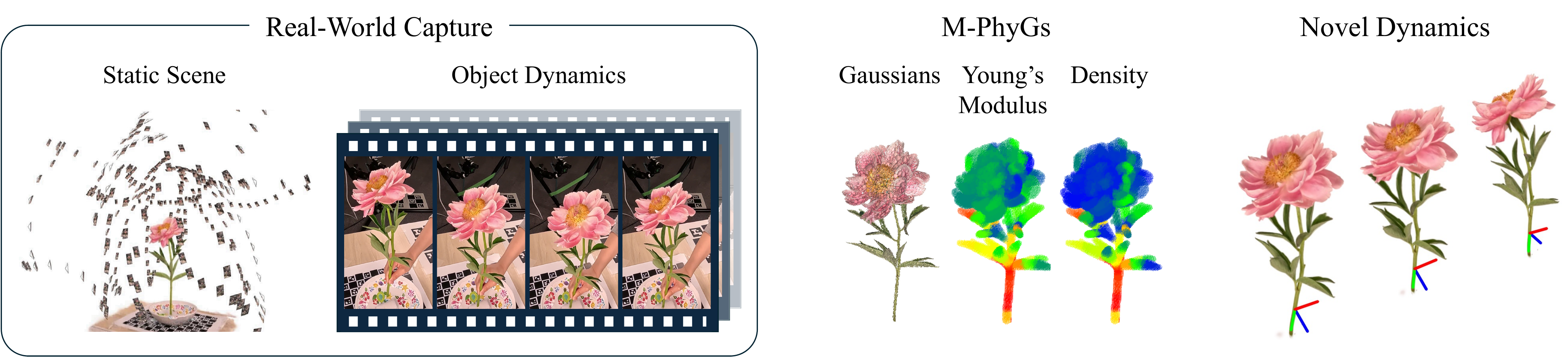}
  \end{center}

  \refstepcounter{figure}\label{fig:opening}
  \noindent\textbf{Figure \thefigure.}
  Multi-material Physical Gaussians (M-PhyGs) recovers the physical material properties such as the Young's modulus of 
  real-world objects consisting of multiple parts of different materials from short videos captured from a sparse set of views. The recovered material parameters can be used to predict how the object would respond to unseen physical interactions.
  \bigskip
}]
\thispagestyle{empty}

\begin{abstract}
  Knowledge of the physical material properties governing the dynamics of a real-world object becomes necessary  to accurately anticipate its response to unseen interactions. Existing methods for estimating such physical material parameters from visual data assume homogeneous single-material objects, pre-learned dynamics, or simplistic topologies. Real-world objects, however, are often complex in material composition and geometry lying outside the realm of these assumptions. In this paper, we particularly focus on flowers as a representative common object. We introduce Multi-material Physical Gaussians (M-PhyGs) to estimate the material composition and parameters of such multi-material complex natural objects from video. From a short video captured in a natural setting, M-PhyGs jointly segments the object into similar materials and recovers their continuum mechanical parameters while accounting for gravity. M-PhyGs achieves this efficiently with newly introduced cascaded 3D and 2D losses, and by leveraging temporal mini-batching. We introduce a dataset, Phlowers, of people interacting with flowers as a novel platform to evaluate the accuracy of this challenging task of multi-material physical parameter estimation. Experimental results on Phlowers dataset demonstrate the accuracy and effectiveness of M-PhyGs and its components. 
\end{abstract}

\section{Introduction}
\label{sec:intro}
\vspace{1.8mm}
Anticipating the dynamic behavior of an object for arbitrary interactions from minimal visual observations can serve an essential role in vision and robotics. Being able to predict, just from a simple visual setup, how an object would behave in response to forces induced through interaction with a human or a robot can enable accurate planning of how to handle the object. In addition to estimating the physical properties for Newton dynamics to capture rigid-body motions, modeling and recovering the physical properties that dictate the dynamics of deformable objects becomes essential. The real world is filled with soft-material objects that can non-rigidly change their shapes as they are simply picked up, carried, and put down. 

Past methods for estimating the underlying parameters or directly learning the dynamics of deformable objects take three distinct approaches. The first assumes single material objects~\cite{li23pacnerf, cai24gic, Cao24neuma, liu2025physflow}, \ie, only recover one set of physical material parameters of a dynamics model (\eg, Material Point Method~\cite{stomakhin13snow,jiang16mpmcourse,hu2018mlsmpm}) for the whole object.
The second pre-learns the dynamics itself (typically with video diffusion) and estimates the physical material parameters with its supervision~\cite{zhang24physdreamer, huang24dreamphysics, liu24physics3d, lin25omniphysgs}, assigns them directly with LLMs~\cite{zhao2024efficient, lin2024phys4dgen}, or trains neural networks on datasets annotated by LLMs~\cite{le2025pixie}.
The third approach models an object as a spring-mass system~\cite{zhong24springgaus, jiang2025phystwin} or a graph neural network~\cite{zhang2024dynamics, shao2025GausSim} of particles (typically 3D Gaussians for Gaussian splatting~\cite{kerbl233Dgaussians}). This inevitably assumes simple topology which can be far from the true internal mechanical topology of the object, which consequently also necessitates diverse motion observations to learn its parameter values.

Natural objects are often made of multiple parts, with distinct boundaries, each made of different materials that exhibit different mechanics seamlessly interacting with each other, leading to complex dynamics as a whole. Think of trees outside the window, and flowers in your garden. They are all composed of different materials such as stems, leaves, and petals. They also have a complex geometry that is difficult to model with simple topology. In this paper, we focus on flowers as daily representative multi-material objects that pose severe challenges to current approaches as their complex material composition fundamentally breaks underlying assumptions, which we also experimentally demonstrate.

A number of challenges underlie the modeling of dynamics of multi-material objects. 
First is the segmentation and estimation of different material segments and parts of the object. Accurate material-wise segmentation cannot be achieved solely from static observations and the physical parameters of each segment cannot be supplied by a pre-learned model, since even objects within the same category have unique material compositions and properties. The interaction of the distinct materials across different parts also add to their rich dynamics. This necessitates an analysis by synthesis approach that combines observations and physics simulation, which gives rise to a slew of difficulties. The 3D geometry of the object can only be recovered in detail with a dense set of views, which is prohibitive for capturing its dynamics. The object whether in motion or in steady state is always in equilibrium with gravity, so gravity needs to be properly accounted for. 
A sizable sequence length of observations of the object dynamics becomes essential for accurate parameter recovery, which causes unstable estimation due to large discrepancies between the predicted and observed as dynamics simulation is inherently sequential. Computational cost also becomes a major obstacle. 

We introduce Multi-material Physics Gaussians (M-PhyGs /\textepsilon m-figz/), a novel multi-material estimation method that overcomes all these challenges. 
M-PhyGs represents the target object with a hybrid representation consisting of 3D Gaussians recovered via 3D Gaussian splatting from a dense view capture of the object in rest state and also dense particles in a regular grid that drive these 3D Gaussians. Physical material properties are assigned to each 3D Gaussian from neighboring grid particles, which are estimated from the dynamics of the object observed from a sparse set of views. M-PhyGs makes four key contributions to enable this multi-material dynamics modeling:  
1) this hybrid represention for dynamics and appearance;
2) joint segmentation and material parameter estimation;
3) inclusion of gravity in the dynamics modeling and material estimation;
4) novel 3D and 2D supervisions and temporal mini-batching for robust and efficient estimation. 

We introduce a first-of-its-kind dataset of human-flower interactions for rigorous comparative analysis of the effectiveness of M-PhyGs. The dataset, which we refer to as Phlowers, captures a person arranging a flower with a sparse set of cameras whose rest shape is densely captured for 3D Gaussian splatting. We conduct extensive experiments on this dataset and evaluate the prediction accuracy for unseen frames. 
The results clearly show that M-PhyGs achieves state-of-the-art accuracy on this challenging task of multi-material object dynamics modeling. All code and data will be made public at publication to catalyze research on this important topic towards embodied vision.

\section{Related Work}
\label{sec:related}

\begin{table*}[t]
    \centering
    \small
    \caption{
    M-PhyGs is the first method to achieve continuum mechanical material parameter estimation for complex, multi-material objects from real videos. Past methods either specialize in synthetic objects or dynamics (not ``Real Dynamics"), single-material objects (not ``Multi-material"),  can only handle simple shapes (not ``Complex Geometry"), or require a large amount of training data (Not ``Data Efficient"). 
    }
    \tabcolsep = 5pt
    \renewcommand{\arraystretch}{1.3}
    \begin{tabularx}{\textwidth}{l|ccccccc}

        & Homogeneous & Diffusion Models & LLMs & Spring-Mass Model & Feed-Forward & GNNs & M-PhyGs\\
        & \footnotesize{\cite{li23pacnerf}, \cite{cai24gic}, \cite{Cao24neuma}, \cite{liu2025physflow}} &
        \footnotesize{\cite{zhang24physdreamer}, \cite{huang24dreamphysics}, \cite{liu24physics3d}, \cite{lin25omniphysgs}} & 
        \footnotesize{\cite{zhao2024efficient}, \cite{lin2024phys4dgen}} &
        \footnotesize{\cite{zhong24springgaus}, \cite{jiang2025phystwin}} &
        \footnotesize{\cite{chen2025vid2sim}, \cite{le2025pixie}} &
        \footnotesize{\cite{zhang2024dynamics}, \cite{shao2025GausSim}} &
        (Ours) \\
        \hline
        Real Dynamics & \cmark & \xmark & \xmark & \cmark & \xmark & \cmark & \cmark \\
        Multi-material & \xmark & \cmark & \cmark & \cmark & \cmark & \cmark & \cmark \\
        Complex Geometry & \cmark & \cmark & \cmark &  & \cmark & \xmark & \cmark \\
        Data Efficient & \cmark & \cmark & \cmark & \cmark & \cmark & \xmark & \cmark 
    \end{tabularx}
    \label{tab:method_comparison} 
\end{table*}

A range of methods has been proposed for modeling the dynamics of deformable objects. \Cref{tab:method_comparison} summarizes these methods with respect to key characteristics.

\vspace{-8pt}
\paragraph{Single Material Objects}
A variety of methods have been introduced for 3D reconstruction of dynamic scenes~\cite{li22neural3dvideo,luiten2023dynamic,yang24deformable,wu24cvpr4dgs,yang2023gs4d}. For better scene understanding and accurate dynamics representations, those that estimate dynamics by leveraging physical priors have attracted attention.

A key approach to this is analysis-by-synthesis, namely estimation of physical properties by minimizing the discrepancy between the results of forward physics simulation and observations~\cite{li23pacnerf, cai24gic, Cao24neuma, liu2025physflow}. Differentiable simulation (\eg, Material Point Method~\cite{stomakhin13snow,jiang16mpmcourse,hu2018mlsmpm}) can be integrated with differentiable photorealistic object representations (\eg, Neural Radiance Fields (NeRFs)~\cite{mildenhall2020nerf} and 3D Gaussian Splatting~\cite{kerbl233Dgaussians}) to exploit the rendering loss for this minimization.

Feed-forward estimation has also been explored, in which a pre-trained feed-forward network directly estimates physical properties from visual observations (often a single image). Chen \etal~\cite{chen2025vid2sim} fine-tune a large video vision transformer to infer the physical properties of an object from its video. Lv \etal~\cite{lv2025physgm} train a U-Net to predict a probability distribution over the physical properties and the 3D Gaussian splatting parameters for a scene. 

These methods, however, assume a single material for the entire object, and cannot be applied to complex real-world objects composed of more than one material. 

\vspace{-8pt}
\paragraph{Learned Materials}
Several methods leverage pre-learned video diffusion models or large language models (LLMs) to model the dynamics of objects composed of multiple materials. Zhang \etal~\cite{zhang24physdreamer} estimate the physical material properties of an object from a video of its dynamics synthesized by a pre-learned video diffusion model. Follow-up works~\cite{huang24dreamphysics, liu24physics3d, lin25omniphysgs} employ Score Distillation Sampling (SDS)~\cite{poole2022dreamfusion} as supervision. Video diffusion models can generate realistic videos, but these videos are not based on laws of physics and cannot represent differences in the dynamics of different objects and compositions. Note that even for the same category, a different object instance would have a different composition of materials and thus dynamics (consider the flowers of fig and carnation). 

Another group of methods assign physical parameters to each material segment based on their semantic descriptions by using LLMs~\cite{zhao2024efficient, lin2024phys4dgen}. These methods can only identify physical material parameters at the scale of object categories, limiting their abilities to represent the vast variations among instances within the same category. These methods also suffer from the inherent ambiguity of identifying homogeneous material segments of an object from static data. 
Feed-forward inference of material properties for each object segment has also been explored. Le \etal~\cite{le2025pixie} train a 3D U-Net to predict a material field from the CLIP feature~\cite{radford2021learning} of each voxel. 

Again, material-wise segmentation from static visual data is inherently ambiguous (\ie, it is near impossible to tell how an object would move without seeing it move). Pre-learned dynamics are also bounded by the observations in the training data which do not reflect those at inference time as real-world objects exhibit diverse compositions. Even if the overall composition can be similar, unless they are exactly the same (at which point there is no point of estimation), a subtle difference (\eg, different size of one segment) can lead to dramatic differences in the overall dynamics. 

\vspace{-8pt}
\paragraph{Graph-Structured Modeling}
Several methods make assumptions on the mechanical structure of the object for their forward dynamics simulation. 
Approaches based on a spring-mass system~\cite{zhong24springgaus, jiang2025phystwin} or graph neural networks~\cite{zhang2024dynamics, shao2025GausSim} assume neighborhood connectivity of particles representing the geometry and material, which usually does not reflect the actual mechanical structure of complex multi-material real-world objects. 
These methods also require many training sequences (\ie, videos capturing dozens of distinct motion types) since the physical constraints only manifest indirectly.

\vspace{-8pt}
\paragraph{Effect of Gravity}
Existing methods 
conduct simulations either in a zero-gravity environment~\cite{liu2025physflow, zhang24physdreamer, huang24dreamphysics, le2025pixie} or adds a constant external force at every object point even though the observation already includes gravitation~\cite{li23pacnerf, cai24gic, Cao24neuma, liu24physics3d, lin25omniphysgs, chen2025vid2sim}. 
The objects we observe or capture, however, are always deformed under the influence of gravity which is already in effect in the observations and the way it manifests depends on the pose of the object. 
Accounting for gravity already imposed in the observation is thus of particular importance.

\section{M-PhyGs}

\begin{figure*}
  \centering
  \includegraphics[width=\linewidth]{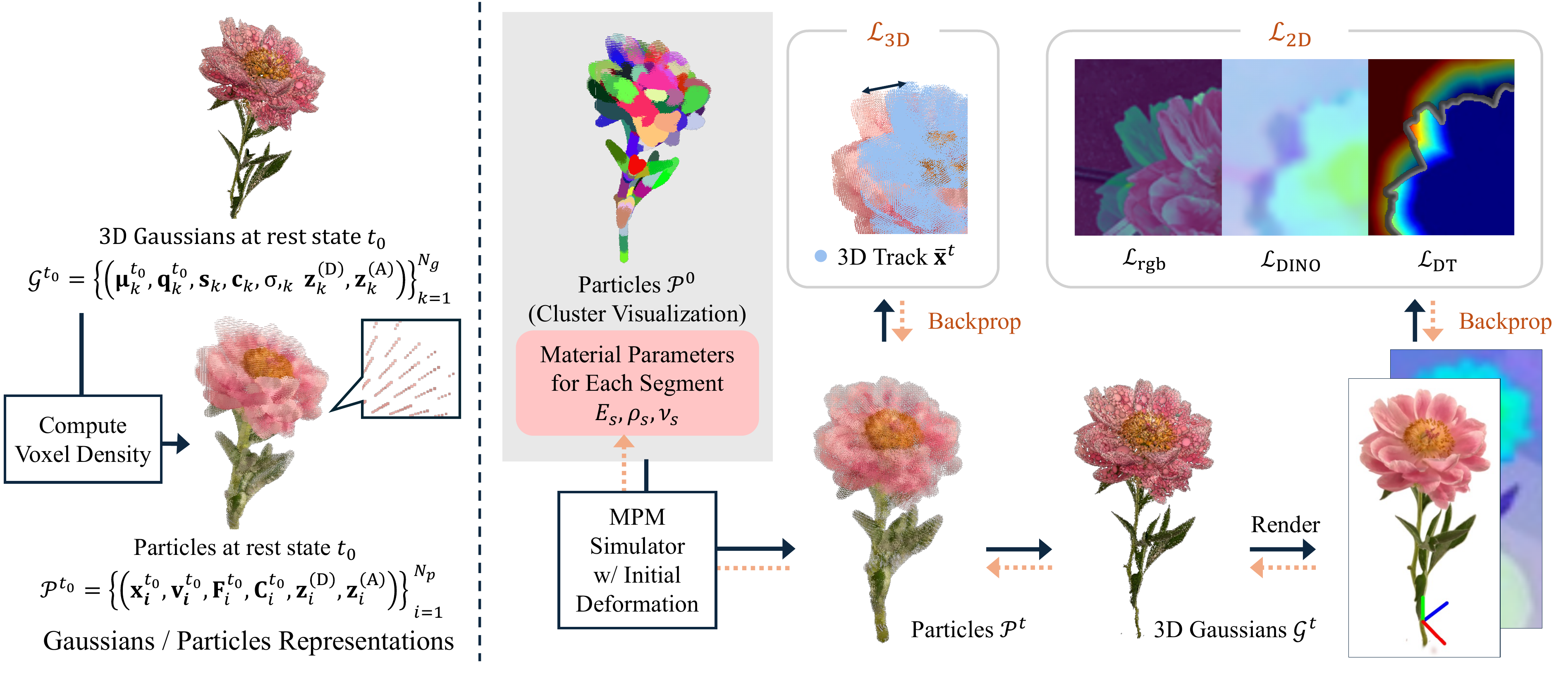}
  \caption{Overview of M-PhyGs. From dense multi-view images of a multi-material deformable object in a static state, we first recover a set of 3D Gaussians and uniformly distribute 3D particles inside the object. From a short video capturing physical interactions with the object captured from a sparse set of views, M-PhyGs estimates the physical material parameters (Young's modulus and density) of these particles which drive the 3D Gaussians. This estimation is achieved by minimization of discrepancies between the predicted and observed dynamics first in 3D geometry by assuming local rigidity and then in the 2D image plane with full non-rigid dynamics.  
  }
  \label{fig:overview}
\end{figure*}

We introduce M-PhyGs, a novel method for estimating the the material properties of multi-material objects from visual observations. \Cref{fig:overview} shows an overview of our method.

\subsection{Photorealistic Dynamics Representation}
\label{sec:3d_representations}

In order to recover the mechanical material parameters of real-world object, we need a representation of the object that can accurately describe its intricate 3D geometry, radiometric appearance, and mechanical dynamics. For this, we adopt 3D Gaussian splatting for the first two and represent their motions due to external and internal forces with 3D particles surrounding them. By simulating the movements of these particles and driving the 3D Gaussians, we can optimize the underlying material parameters so that visual observations captured in the video can be explained. 

The 3D Gaussians can be recovered from a dense-view capture of the object after it settles in rest shape. This can be naturally achieved by scanning an object after (or before) dynamic interaction with it, such as after putting down a deformable object on the desk. The dynamics can in turn be observed from a handful of cameras before (or after) that dense capture, for instance when the object is moved around in a person's hand. As such, both the 3D Gaussians, 3D particles, and their dynamics observation can be extracted from a single sequence of human-object interaction.

Let us denote the set of 3D Gaussians recovered from the rest shape dense capture with $\mathcal{G}$. When the object is moving in the video, these Gaussians can be indexed with time $t$ 
\begin{equation}
    \mathcal{G}^{t} = \left\{ \left(\bm{\mu}_{k}^{t}, \mathbf{q}_{k}^{t}, \mathbf{s}_{k}, \mathbf{c}_{k}, \sigma_{k}, \mathbf{z}_{k}^{\left(\mathrm{D}\right)}, \mathbf{z}_k^{\left(\mathrm{A}\right)} \right) \right\} _{k=1}^{N_g} \,,
\end{equation}
where $\bm{\mu}_{k}^{t}$ and $\mathbf{q}_{k}^{t}$ are the position and rotation (quaternion) of the $k$-th Gaussian at time $t$, respectively. $N_g$ is the number of Gaussians. $\mathbf{s}_{k}, \mathbf{c}_{k}$, and $\sigma_{k}$ are the 3D scale, RGB color, and opacity parameters, respectively, which we model as being time-invariant (\ie, fixed-sized Gaussians and Lambertian surfaces). We use the frame number for $t$. During the reconstruction of 3D Gaussians, we also recover a DINO~\cite{simeoni2025dinov3} feature vector $\mathbf{z}_k^{\left( \mathrm{D} \right)}$ and affinity feature vector $\mathbf{z}_k^{\left( \mathrm{A} \right)}$ for each Gaussian~\cite{kerr2024rsrd,cmk2024garfield}. These features are optimized with 2D feature maps extracted from the multi-view images similar to the optimization of Gaussian colors $\mathbf{c}_k$ with RGB images.

3D Gaussian splatting optimizes the Gaussians to represent the outer surface of an object and is not suitable for representing the dynamics of the object. We inject a set of 3D particles $\mathcal{P}$ in the volume subtended by the 3D Gaussians, simulate the dynamics of these particles, and move the Gaussians based on them. Particle states at time $t$ are 
\begin{equation}
    \mathcal{P}^{t} = \left\{ \left(\mathbf{x}_i^{t}, \mathbf{v}_{i}^{t}, \mathbf{F}_{i}^{t}, \mathbf{C}_{i}^{t}, \mathbf{z}_{i}^{\left(\mathrm{D}\right)}, \mathbf{z}_{i}^{\left(\mathrm{A}\right)} \right)\right\}_{i=1}^{N_p} \,,
\end{equation}
where $\mathbf{x}^t_i$, $\mathbf{v}^t_i$, $\mathbf{F}^t_i \equiv \frac{\partial\mathbf{x}_i^t}{\partial \mathbf{x}_i^0}$ and $\mathbf{C}^t_i \equiv \frac{\partial \mathbf{v}_i^t}{\partial \mathbf{x}_i^t}$ are the 3D location, velocity, deformation gradient, and affine velocity of the $i$-th particle at time $t$, respectively, and $N_p$ is the number of particles. $\mathbf{z}_{i}^{\left(\mathrm{D}\right)}$ and $\mathbf{z}_{i}^{\left(\mathrm{A}\right)}$ are the DINO feature and affinity feature, respectively, for each particle which are assigned from the nearest Gaussians of the rest shape.

Particles at rest state are uniformly distributed in the object volume, which is defined by the voxel density derived from the distribution of 3D Gaussians. 
Since the voxel density is continuous, its boundary is not apparent. We first sample the point cloud of the volume defined by a loose threshold on the density values, and optimize an additional parameter that controls how far outside points from the boundary are included in the simulation. 

Material parameters, namely density $\rho_i$, Young's modulus $E_i$, and Poisson's ratio $\nu_i$ are also assigned to the particles. We assume constant Poisson's ratio $\nu_i$ as its possible range is small, and estimate $E_i $ and $\rho_i$ through the minimization. 
M-PhyGs simulates the motion of particles with material parameters of each particle and then drives nearest neighboring Gaussians accordingly.

\subsection{Forward Dynamics with Gravity}
\label{sec:dynamics_representation}

M-PhyGs simulates the dynamics of the object with the estimated material parameters with a continuum mechanics simulator based on the Moving Least Squares Material Point Method (MLS-MPM)~\cite{hu2018mlsmpm}. MLS-MPM takes physical states of particles at a given time as inputs, and outputs those of the next timestep based on the physical material parameters of each particle. 
Interaction with objects external to the target object (\eg, a human hand) can also be simulated by adding boundary conditions on particle velocities. The motion of the contact point and the initial position of particles are estimated from the video.
Please see the appendix for details.

All objects in the world, including the captured one, are under the influence of gravity. Surprisingly, past methods ignore this fact and simulate the dynamics under zero gravity~\cite{zhang24physdreamer, liu2025physflow, huang24dreamphysics, le2025pixie} or additive constant gravity without considering the gravity-induced deformation of an object already included in its initial state (\ie, initial deformation gradient is set to identity)~\cite{li23pacnerf, cai24gic, Cao24neuma, liu24physics3d, lin25omniphysgs, chen2025vid2sim}. 

Properly accounting for gravity when estimating materials is essential for the subsequent dynamics simulation. 
For this, effects of gravity on the rest shape needs to be estimated. The deformation gradient, however, expressed as a $3 \times 3$ matrix, must satisfy several constraints (\eg, non-degenerate and positive definite), which makes this estimation challenging.

\begin{figure}[t]
  \centering
  \includegraphics[width=\linewidth]{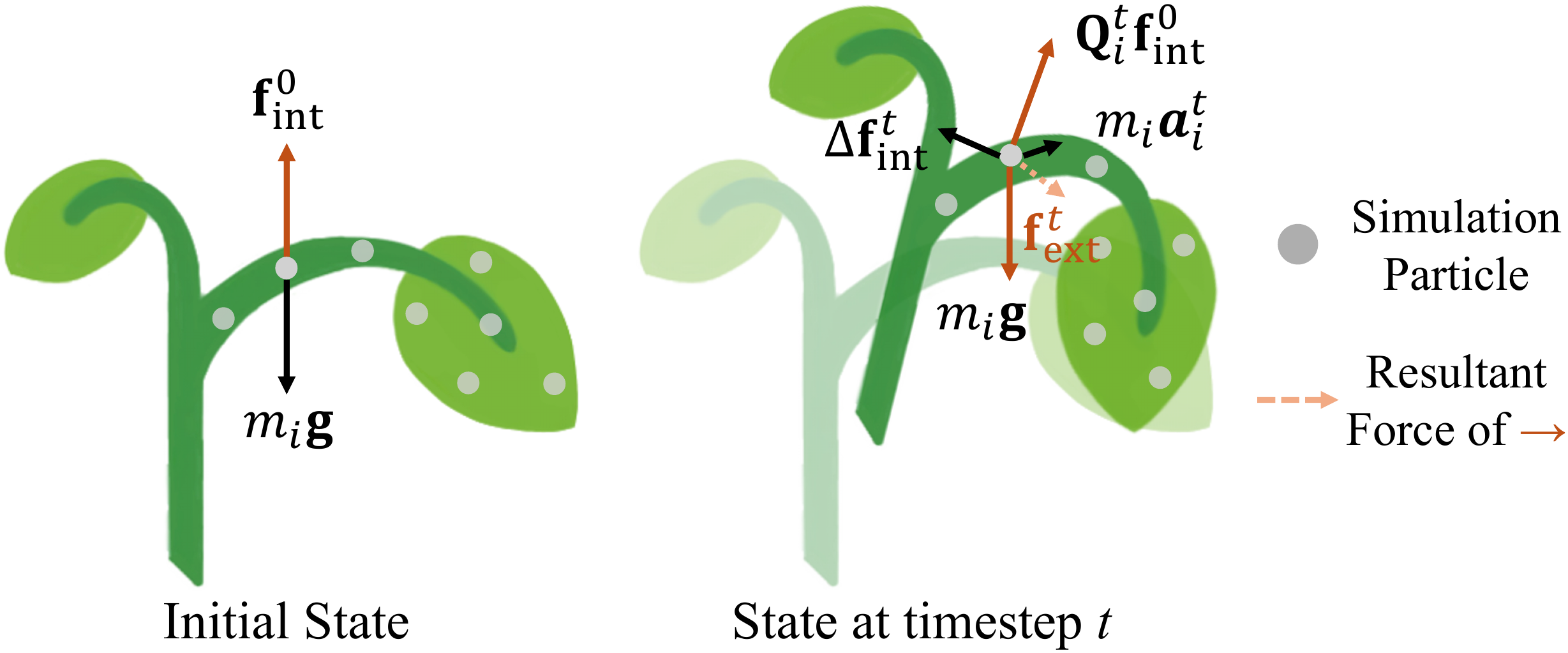}
  \caption{M-PhyGs accounts for gravity in the forward dynamics computation by adding a rotated initial internal force to counter the gravitational force at rest shape.}
  \label{fig:gravity}
\end{figure}

M-PhyGs instead estimates initial internal force per particle $\mathbf{f}_\mathrm{int}$ and accommodates it in the subsequent dynamics simulation as an external force. As depicted in \cref{fig:gravity}, in the initial state, assuming that a particle acceleration is sufficiently small (\ie, it is in rest shape), we can compute $\mathbf{f}_\mathrm{int}^0$ from the gravitational acceleration $\mathbf{g}$
\begin{equation}
    \mathbf{f}_\mathrm{int}^{0} = - m_i \mathbf{g} \,,
\end{equation}
where $m_i$ is the particle mass.
In subsequent MPM simulation, the effect of gravity and initial deformation can be approximated by adding 
\begin{equation}
    \mathbf{f}_\mathrm{ext} = m_i \mathbf{g} + \mathbf{Q}_i^t \mathbf{f}_\mathrm{int}^0 \,,
    \label{eq:gravity_formulation}
\end{equation}
where $\mathbf{Q}_i^t$ is a rotation matrix that represents the relative rotation between timestep $0$ and $t$, which is computed by singular value decomposition of the deformation gradient $\mathbf{F}_i^t$. As a result,  
\begin{equation}
    \mathbf{f}_\mathrm{ext} + m_i \mathbf{a}_i^{t} + \Delta \mathbf{f}_\mathrm{int}^{t} = 0
\end{equation}
holds true at any given time, where $\mathbf{a}_i^t$ and $\Delta \mathbf{f}_\mathrm{int}^{t}$ denote the acceleration and the change of internal force from the initial state of $i$-th particle at time $t$, respectively.

\subsection{Multi-Material Estimation}
\label{sec:material_parameter_estimation}

\begin{figure}[t]
  \centering
  \includegraphics[keepaspectratio, width=\linewidth]{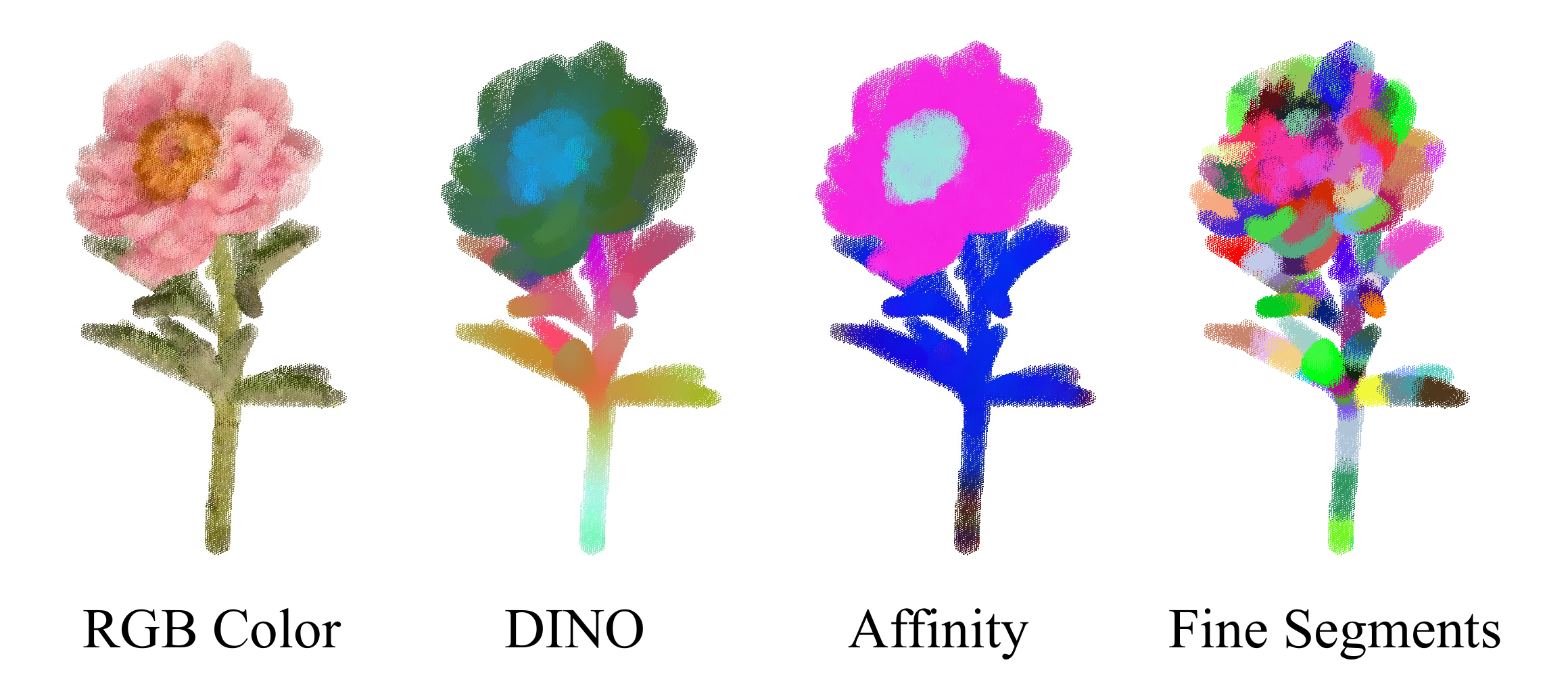}
  \caption{M-PhyGs leverages DINO~\cite{simeoni2025dinov3} and GARField (affinity)~\cite{cmk2024garfield} features assigned to each 3D particle for initial fine-grained material segmentation. It estimates per-segment physical material parameters and encourages further merging of the segments with a material grouping loss. 
  }
  \label{fig:grouping}
\end{figure}

The large number of free parameters makes per-particle material estimation extremely challenging. 

\vspace{-8pt}
\paragraph{Joint Segmentation}
We resolve this by jointly segmenting the object into segments of homogeneous material and by estimating the material parameters for each. 
M-PhyGs achieves this by leveraging the visual features~\cite{simeoni2025dinov3,cmk2024garfield} recovered for both the Gaussians and physical particles, since parts with similar appearance within the object tend to have similar physical properties.
\Cref{fig:grouping} shows how the Gaussians and the physical particles are segmented from these features.
Hyperparameters of the segmentation algorithm are adjusted to oversegment for subsequent grouping of the materials.

M-PhyGs estimates the material parameters, $E_s$ and $\rho_s$, of each material segment $s$.
M-PhyGs further consolidates the material segments with a material similarity loss based on the DINO features ~\cite{simeoni2025dinov3}: 
\begin{equation}
    \mathcal{L}_\mathrm{s} = \frac{1}{N_p} \sum_i \sum_{j \in \mathcal{N}_i} \sum_{a \in \{\rho, E\}} w_{ij} \left\lVert \log\left(a_i\right) - \log\left(a_j\right) \right\rVert^2 \,,
\end{equation}
where $\rho_i$ and $E_i$ here are the density and Young's modulus of the $i$-th particle's segment, respectively, and $\mathcal{N}_i$ is a set of indices of neighboring particles of the $i$-th particle in the DINO feature space. We use 20 neighbors in our experiment. The weight $w_{ij}$ is
\begin{equation}
    w_{ij} = \frac{\exp\left( - \alpha \left\lVert \mathbf{z}_i^{\left(\mathrm{D}\right)} - \mathbf{z}_j^{\left(\mathrm{D}\right)} \right\rVert^2_2 \right) + \varepsilon}{\sum_{j \in \mathcal{N}_i} \left( \exp\left( - \alpha \left\lVert \mathbf{z}_i^{\left(\mathrm{D}\right)} - \mathbf{z}_j^{\left(\mathrm{D}\right)} \right\rVert^2_2 \right) + \varepsilon \right)} \,.
\end{equation}
In practice, we set $\alpha$ to 20 and $\varepsilon$ to $1 \times 10^{-9}$.

We also discourage the range of per-segment physical material parameter values from becoming too large with a material variance loss
\begin{equation}
    \mathcal{L}_\mathrm{v} = \mathrm{var}\left(\log\left(E_i\right)\right) + \mathrm{var}\left(\log\left(\rho_i\right)\right) \,,
\end{equation}
where $\mathrm{var}\left(x_i\right)$ is the variance of $x_i$.
The material grouping loss is a weighted sum of the material similarity loss and the material variance loss
\begin{equation}
    \mathcal{L}_\mathrm{g} = \mathcal{L}_\mathrm{s} + w_\mathrm{v} \mathcal{L}_\mathrm{v} \,,
\end{equation}
where $w_\mathrm{v}=1$ in practice. 

\vspace{-8pt}
\paragraph{3D and 2D Supervisions}

Accurate material estimation requires observation of meaningful deformation of the object in a video capture of sufficient length. As the continuum mechanical simulation is necessarily sequential, this leads to large discrepancies between the predicted and observed dynamics, especially in the early stages of the analysis-by-synthesis loop, which leads to divergence of the parameter estimation. 
M-PhyGs overcomes this by first performing coarse optimization using 3D Gaussian tracking with local rigidity constraints.
Material properties are then optimized with the actual observed 2D ground truth. As such, it is a coarse-to-fine estimation realized with cascaded optimization in geometry and photometry and from rigid to non-rigid in terms of the object dynamics. 

For the coarse 3D optimization, we supervise M-PhyGs with tracked 3D Gaussians using Dynamic 3D Gaussians~\cite{luiten2023dynamic} which assumes local rigidity. 
The material parameters are estimated by minimizing the discrepancy between the 3D tracks ${\bar{\mathbf{x}}_i^{t}}$ and those simulated from the estimates
\begin{equation}
    \mathcal{L}_\mathrm{3D} = \sum_{t=1}^T \sum_i \left\lVert\hat{\mathbf{x}}_i^t\left(\theta \right) - \bar{\mathbf{x}}_i^t\right\rVert^2 \,,
\end{equation}
where $\hat{\mathbf{x}}_i^t\left(\theta \right)$ is the location of the simulated particles computed from a set of material parameters $\theta$. 

Once the optimization with $\mathcal{L}_\mathrm{3D}$ converges, M-PhyGs refines the material parameters to capture the full non-rigid dynamics by minimizing a loss in the 2D image plane. The key idea here is to leverage discrepancies in image features and the object boundaries between the prediction and observation. The DINO feature loss $\mathcal{L}_\mathrm{DINO}$ is an L1 loss on the rendered and ground-truth (observed) feature maps evaluated at multiple resolutions. The object boundary loss is imposed with a distance transform 
\begin{equation}
    \mathcal{L}_\mathrm{DT} = \frac{1}{N_c} \sum_{c}^{N_c} \left(\frac{1}{N_g} \sum_k^{N_g} \mathcal{D}_{c} \left( \pi_c \left( \bm{\mu}_k \right)\right)\right) \,,
\end{equation}
where $\pi_c$ is a projection function of $c$-th view which takes the position of the $k$-th Gaussian, $\bm{\mu}_k$, as an input, $\mathcal{D}_{c}$ returns a precomputed distance map value of the $c$-th view at the input pixel, and $N_c$ is the number of viewpoints. Distance maps are calculated in advance from a SAM2~\cite{ravi2024sam2} mask of the foreground object region.

These losses encourage the alignment of the silhouette and image-feature distributions, acting as a correction that is more global than the RGB loss, $\mathcal{L}_\mathrm{rgb}$, which is defined as a weighted sum of an L1 term of the RGB images at multiscale resolution and a D-SSIM term. The complete 2D supervision is
\begin{equation}
    \mathcal{L}_\mathrm{2D} = w_\mathrm{rgb} \mathcal{L}_\mathrm{rgb} + w_\mathrm{DINO} \mathcal{L}_\mathrm{DINO} + w_\mathrm{DT} \mathcal{L}_\mathrm{DT} ,
\end{equation}
with $w_\mathrm{rgb} = 0.1$, $w_\mathrm{DINO} = 0.1$, and $w_\mathrm{DT} = 1 \times 10^{-3}$ in our experiment.

\vspace{-8pt}
\paragraph{Temporal Mini-Batching}

The MPM simulator calculates particle states sequentially causing errors to accumulate over time. Its computational cost is also a major issue.
We stabilize and accelerate the optimization in M-PhyGs by splitting the video frames into temporal mini-batches. 
The temporal segregation helps limit the simulation error accumulation, and at the same time, enables parallelization of optimization. M-PhyGs first computes the initial position $\mathbf{x}_i^t$, velocity $\mathbf{v}_i^t$, and acceleration $\mathbf{a}_i^t$ of the particles for each temporal mini-batch from the 3D tracks $\bar{\mathbf{x}}_i^t$. These approximated initial physical particle states are then used for the parallel temporal batch-wise simulation. The losses are back-propagated from all mini-batches to a shared set of physical material parameters for each material segment.

\begin{figure*}
  \centering
  \includegraphics[width=\linewidth]{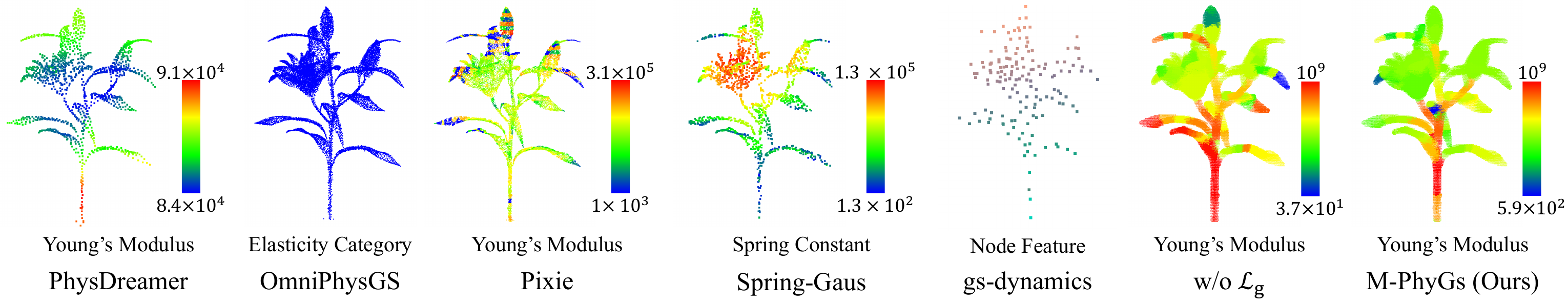}
  \caption{Estimated physical material parameters of our method and existing methods~\cite{lin25omniphysgs,zhang24physdreamer,le2025pixie,zhong24springgaus,zhang2024dynamics}. 
  For OmniPhysGS~\cite{lin25omniphysgs}, the constitutive model (\ie, probability of whether the particle is elastic or not), and for gs-dynamics~\cite{zhang2024dynamics}, the node features are shown, respectively.
  The estimated per-segment material parameters of M-PhyGs form clusters that roughly align with the different object parts. 
  }

  \label{fig:result_mat_params}
\end{figure*}

\begin{table*}[t]
    \centering
    \small
    \caption{
      Quantitative accuracy comparison of dynamics prediction using estimated material. M-PhyGs achieves state-of-the-art accuracy in predicting unseen dynamics and estimating material parameters of complex multi-material flowers. 
    }
    
    \begin{tabular}{l|rrr|rrr}
        & \multicolumn{3}{c|}{In-Sequence} & \multicolumn{3}{c}{Cross-Sequence} \\
        & \multicolumn{1}{c}{{PSNR ($\uparrow$)}} & \multicolumn{1}{c}{{IoU ($\uparrow$)}} & \multicolumn{1}{c|}{{CD ($\downarrow$)}} 
        & \multicolumn{1}{c}{{PSNR ($\uparrow$)}} & \multicolumn{1}{c}{{IoU ($\uparrow$)}} & \multicolumn{1}{c}{{CD ($\downarrow$)}} \\ \hline

        {PhysDreamer~\cite{zhang24physdreamer}} & 16.00 & 31.88\% & 52.6 px & 15.66 & 25.24\% & 16.7 px \\
        {OmniPhysGS~\cite{lin25omniphysgs}} & 15.31 & 8.17\% & 163.6 px & 15.39 & 15.18\% & 129.3 px \\
        {Pixie~\cite{le2025pixie}} & 15.63 & 22.52\% & 72.0 px & 16.45 & 31.48\% & 58.55 px \\
        {Spring-Gaus~\cite{zhong24springgaus}} & 15.61 & 6.58\% & 174.7 px & 15.54 & 9.03\% & 169.8 px \\
        {gs-dynamics~\cite{zhang2024dynamics}} & 16.13 & 34.28\% & 40.6 px & 16.31 & 29.69\% & 65.6 px \\
        {GIC~\cite{cai24gic}} & 15.78 & 9.46\% & 169.6 px & 15.85 & 6.21\% & 208.1 px \\
        \textbf{M-PhyGs (Ours)} & \textbf{18.49} & \textbf{70.58\%} & \textbf{3.3 px} & \textbf{17.99} & \textbf{65.13\%} & \textbf{4.9 px} \\
         
    \end{tabular}

    \label{tab:result_future_prediction} 
\end{table*}

\begin{figure*}
  \centering

  \begin{subfigure}{\textwidth}
    \centering
    \includegraphics[width=\textwidth, keepaspectratio]{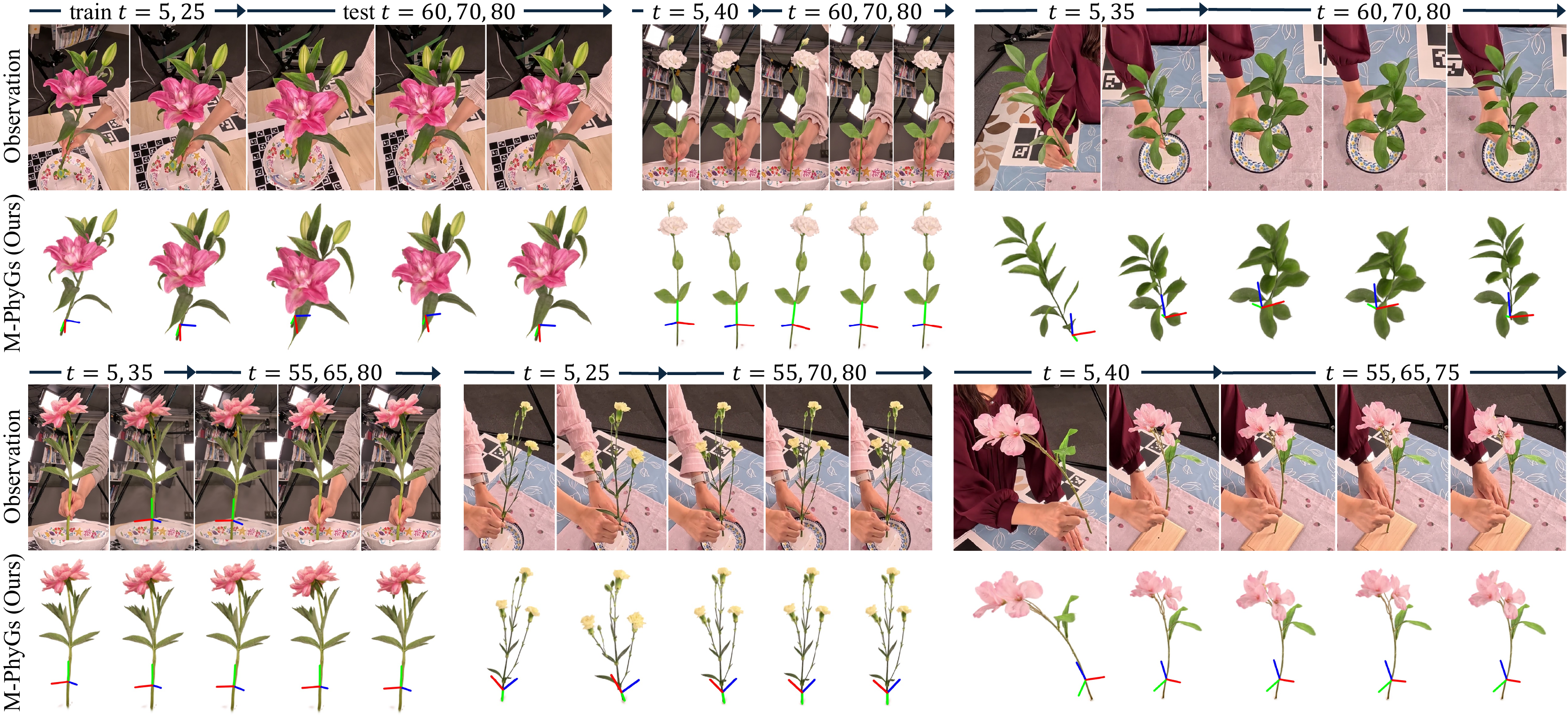}
    \caption{Results of M-PhyGs (Ours) and ground-truth observations.}
    \label{fig:predict_ours}
  \end{subfigure}

  \par\medskip

  \begin{subfigure}{\textwidth}
    \centering
    \includegraphics[width=\textwidth, keepaspectratio]{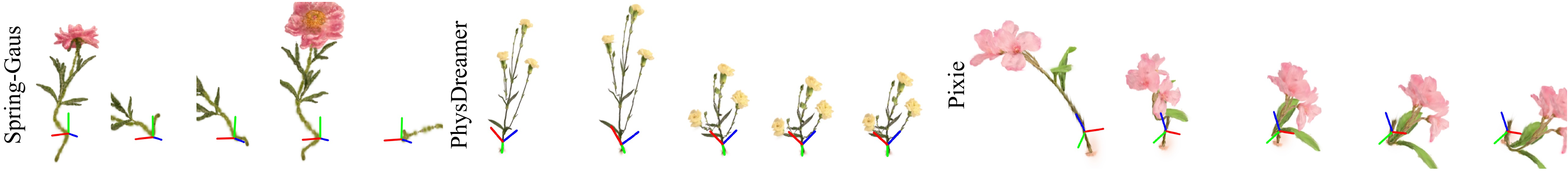}
    \caption{Results of existing methods.}
    \label{fig:predict_comparison}
  \end{subfigure}

  \caption{
    Unseen (in-sequence) dynamics predicted with estimated physical material parameters. For each object, the first two show samples of  training frames and the last three show samples of predicted frames. M-PhyGs predicts the complex motion of each flower which aligns with the actual held out observations. In contrast, existing methods fatally diverge from the true motion often completely collapsing due to erroneous material estimates. 
  }
  \label{fig:result_future_prediction}
\end{figure*}

\begin{figure*}
  \centering

  \begin{subfigure}{\textwidth}
    \centering
    \includegraphics[width=\textwidth, keepaspectratio]{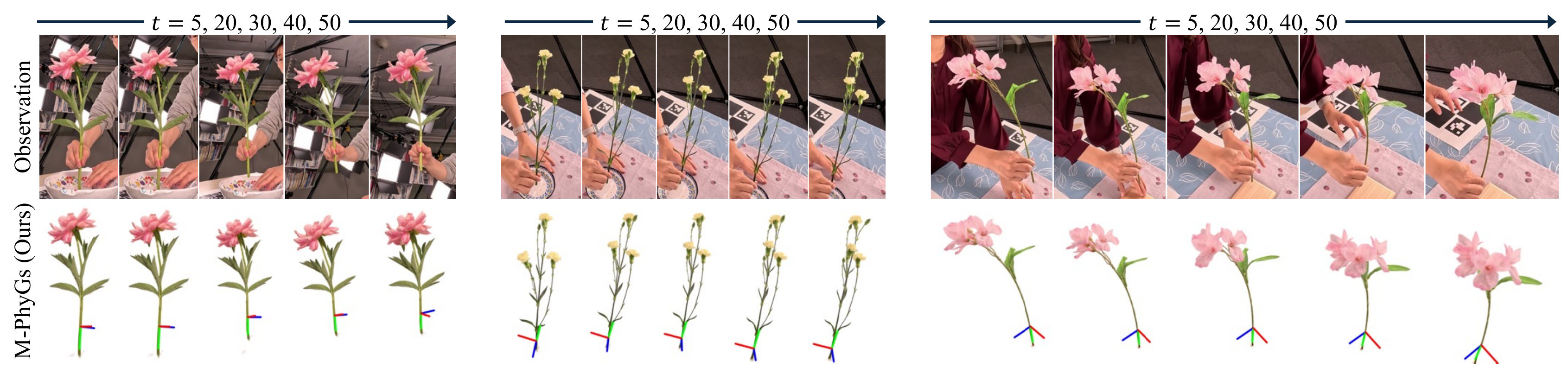}
    \caption{Results of M-PhyGs (Ours) and ground-truth observations.}
    \label{fig:predict_ours_cross}
  \end{subfigure}

  \par\medskip

  \begin{subfigure}{\textwidth}
    \centering
    \includegraphics[width=\textwidth, keepaspectratio]{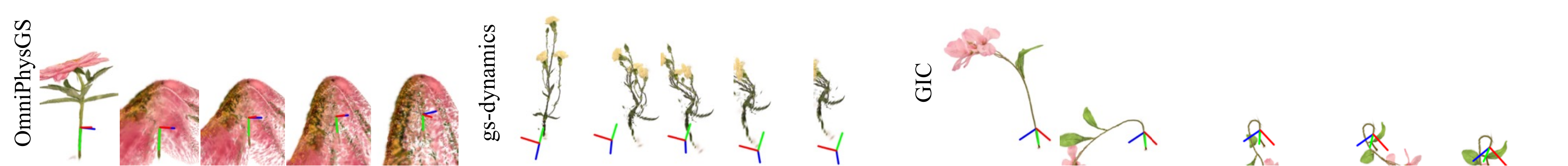}
    \caption{Results of existing methods.}
    \label{fig:predict_comparison_cross}
  \end{subfigure}

  \caption{
    Cross-sequence dynamics predicted with estimated physical material parameters. M-PhyGs successfully predicts the complex motion of each flower even in this fully unseen interactions.
  }
  \label{fig:result_cross_prediction}
\end{figure*}

\section{Experimental Results}

We experimentally validate the effectiveness of our method on newly captured real data, as none of the past public datasets capture multi-material objects. 

\vspace{-8pt}
\paragraph{Phlowers Dataset} We introduce a novel dataset, which we refer to as Phlowers dataset (physics of flowers) specifically focused on real flowers as a representative and challenging but natural multi-material object. Phlowers consists of real multi-view videos of 10 flowers. For each flower, we captured videos from 5 different viewpoints as a person inserts the flower into a flower frog. Each video contains at least 100 frames. The intrinsic and extrinsic camera parameters are estimated with COLMAP~\cite{schoenberger2016sfm,schoenberger2016mvs} together with the dense view capture of the static scene. The coordinate scales and rotations are aligned using ChArUco boards, and the videos are synchronized using time code. 

\vspace{-8pt}
\paragraph{Evaluation Metric} It is near-impossible to measure the ground-truth physical material parameters of real-world objects. 
We quantitatively evaluate the accuracy of the material parameter estimates by their ability to accurately predict the dynamics of the object for unseen interactions.
For each set of multi-view videos, we estimate the segmentation and material parameters of the object from the first 50 frames and predict the physical states and corresponding rendered views for the 51st to 80th frames. We evaluate the accuracy of the predicted (rendered) videos with PSNR, 2D IoU, and 2D chamfer distance (CD) using masks annotated by SAM2~\cite{ravi2024sam2} as pseudo ground truth.

\vspace{-8pt}
\paragraph{Baseline Methods} We compare the accuracy of our M-PhyGs with a variety of methods that are representative of the distinct approaches explored in the past. 
PhysDreamer~\cite{zhang24physdreamer} generates videos from a single image input by a diffusion model and estimates physical material parameters from the generated videos. OmniPhysGS~\cite{lin25omniphysgs} optimizes materials by SDS loss~\cite{poole2022dreamfusion}. 
Pixie~\cite{le2025pixie} trains a feed-forward network to estimate these properties.
Spring-Gaus~\cite{zhong24springgaus} and gs-dynamics~\cite{zhang2024dynamics}
are methods based on particle connectivity, such as a spring-mass model or a graph neural network. 
GIC~\cite{cai24gic} exploits an MPM simulator but assumes homogeneous material.

\subsection{Material Parameter Estimation}

\Cref{fig:result_mat_params} visualizes physical material parameters estimated by M-PhyGs and past methods~\cite{zhong24springgaus, lin25omniphysgs, zhang24physdreamer, zhang2024dynamics}. 
Past methods struggle with multi-material objects as they directly optimize material parameters for each particle or strongly impose spatial smoothness.
In contrast, the estimated per-segment material parameters of our method form physically plausible clusters that align well with the actual object part decomposition. This demonstrates the effectiveness of our method in material segmentation and parameter estimation.

\vspace{-6pt}
\paragraph{Dynamics Prediction}

\Cref{fig:result_future_prediction,fig:result_cross_prediction} and \cref{tab:result_future_prediction} show qualitative and quantitative results of dynamics prediction for unseen interactions using the parameter estimates. 
In-sequence results show the dynamics prediction for the training data and its subsequent sequence, and cross-sequence dynamics indicate a sequence distinct from the training data.
The results by PhysDreamer~\cite{zhang24physdreamer} 
, OmniPhysGS~\cite{lin25omniphysgs}
, and Pixie~\cite{le2025pixie}
are inconsistent with the ground-truth real-world flower motion, often completely collapsing, which shows the difficulty of learning a universal material prior that links the dynamics to appearance. 
Methods that assume simplistic mechanical topologies, \ie, Spring-Gaus~\cite{zhong24springgaus} and gs-dynamics~\cite{zhang2024dynamics},
struggle with the complex structure of multi-material objects. 

In contrast, our method successfully predicts the full movements of the target based on the recovered materials, which demonstrates the accuracy of the method. 
Please see the appendix for more qualitative results.

\subsection{Ablation Studies}

\begin{figure}
  \centering
  \includegraphics[width=\linewidth]{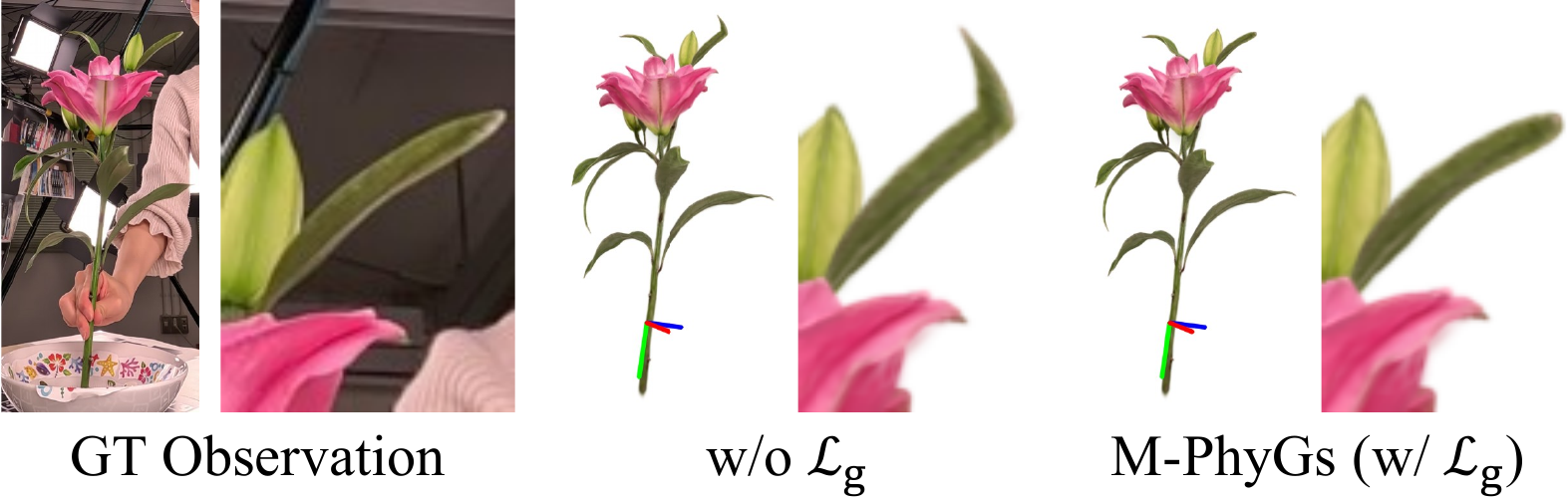}
  \caption{The material grouping loss $\mathcal{L}_\mathrm{g}$ encourages better segmentation of material regions and leads to more accurate dynamics prediction.}
  \label{fig:result_ablation_grouping}
\end{figure}

\begin{table}[t]
    \centering
    \small
    \caption{
      Quantitative results of ablation studies.
      ``w/o TMB'' denotes w/o temporal mini-batching. 
      Every component of M-PhyGs improves the prediction and material estimation accuracy.
    }
    
    \begin{tabular}{l|rrrrrr}
        & \multicolumn{1}{c}{{PSNR ($\uparrow$)}} & \multicolumn{1}{c}{{IoU ($\uparrow$)}} & \multicolumn{1}{c}{{CD ($\downarrow$)}} \\ \hline
         
         w/o $\mathcal{L}_\mathrm{2D}$ & 18.08 & 76.53\% & 2.1 px \\
         w/o $\mathcal{L}_\mathrm{DT}$ & 18.16 & 77.25\% & 2.1 px & \\
         w/o $\mathcal{L}_\mathrm{g}$ & 18.20 & 77.16\% & 2.3 px \\
         w/o TMB & 17.68 & 71.73\% &  4.2 px\\
         w/o $\mathbf{f}_\mathrm{int}^0$ & 18.13 & 75.82\% & 2.3 px\\
         \textbf{M-PhyGs (Ours)} & \textbf{18.23} & \textbf{77.83\%} & \textbf{1.9 px} \\
        
    \end{tabular}
    \label{tab:result_ablation} 
\end{table}

We conduct ablation studies to study the effectiveness of 
the 2D loss ($\mathcal{L}_\mathrm{2D}$), the object boundary loss ($\mathcal{L}_\mathrm{DT}$), the material grouping loss ($\mathcal{L}_\mathrm{g}$), and the temporal mini-batching. We also compare our method with its own variant that ignores the initial internal force $\mathbf{f}_\mathrm{int}^0$ in \cref{eq:gravity_formulation} and applies a constant gravity force as the external force.
We use 4 flowers in Phlowers dataset for this evaluation.
\Cref{tab:result_ablation} shows quantitative results. The results show that the proposed components improve the accuracy of the dynamics prediction. 
\Cref{fig:result_mat_params,fig:result_ablation_grouping} show qualitative results of M-PhyGs and M-PhyGs w/o the grouping loss. The grouping loss encourages better segmentation of material regions and leads to more accurate dynamics prediction.

\section{Conclusion}
We introduced M-PhyGs, a novel method for modeling the complex dynamics of natural multi-material objects. 
We focused on flowers and introduced Phlowers dataset to validate the effectiveness of M-PhyGs and its advantages over past methods. We believe M-PhyGs can play a role in endowing vision with physical embodiment and Phlowers can serve as a sound platform for further studies on this challenging task of visual dynamics modeling of natural non-rigid objects.

Although our method can deal with complex real-world multi-material objects, some limitations still remain. First, we rely on an off-the-shelf 3D tracking method~\cite{luiten2023dynamic} to obtain motion of the contact point and initial estimates of the physical particle states. As it assumes local rigidity of the object, it can fail on objects with large deformation and fail to bootstrap the estimation. Second, as the material grouping loss is based on DINO features (\ie, object appearance), it would be unsuitable when the physical material properties of internal regions are much different from those of the object surfaces. We plan to tackle these in future work.

\paragraph*{Acknowledgement}
The authors thank Chung Min Kim, Justin Kerr, and Angjoo Kanazawa for their insightful input and discussions. 
This work was in part supported by
JSPS KAKENHI 
21H04893, 
and JST JPMJAP2305. 

\appendix

\section{Object Boundary Estimation}
Since the voxel density computed from the 3D Gaussians is continuous, it is non-trivial to accurately estimate the object boundary from the Gaussians. M-PhyGs resolves this by refining an estimate of the object boundary during the material parameter estimation. When M-PhyGs samples uniformly distributed 3D particles from the object volume defined by a density threshold, it also samples particles from a narrow band enclosing the object boundary. For each particle, we compute a signed distance $d_i$ from the tentative object boundary and determine the particle volume $V_i$ for the MPM simulation accordingly:
\begin{equation}
    V_i = V_g \left\{\operatorname{sigmoid}\left(-\beta \left(d_i + o_{s_i} \right)\right)+\epsilon\right\} \,,
\end{equation}
where $V_g$ is the particle volume computed from the particle spacing, $s_i$ is the material segment to which the $i$-th particle belongs, and $o_s$ is an optimizable boundary offset for each segment $s$. We set $\beta$ to $5 \times 10^{3}$ and $\epsilon$ to $10^{-4}$. This ensures that particles outside the estimated object boundary have almost zero volume and do not affect the MPM simulation.

\section{Incorporating Contact-Point Motion}
\label{sec:contact_point_motion}

M-PhyGs recovers the 6D motion of the contact point from 2D annotations and the particle 3D tracks $\bar{\mathbf{x}}_i^{t}$ used in the computation of the 3D loss $\mathcal{L}_\mathrm{3D}$. It first recovers coarse estimates of the contact point location $\tilde{\mathbf{x}}_c^t$ for each frame $t$ by triangulating manually annotated 2D locations. It then recovers refined per-frame contact point locations $\hat{\mathbf{x}}_c^t$ and per-frame contact point rotations (quaternions) $\hat{\mathbf{q}}_c^t$ by minimizing
\begin{equation}
    \mathcal{L}_\mathrm{c} = \mathcal{L}_\mathrm{a} + \mathcal{L}_\mathrm{t} + \lambda_\mathrm{ps} \mathcal{L}_\mathrm{ps} + \lambda_\mathrm{rs} \mathcal{L}_\mathrm{rs} \,,
\end{equation}
where $\mathcal{L}_\mathrm{a}$ is an annotation loss
\begin{equation}
    \mathcal{L}_\mathrm{a} = \sum_t \left\lVert \hat{\mathbf{x}}_c^t - \tilde{\mathbf{x}}_c^t \right\rVert^2 \,,
\end{equation}
$\mathcal{L}_\mathrm{t}$ is a tracking loss
\begin{equation}
    \mathcal{L}_\mathrm{t} = \sum_t \sum_{i \in \mathcal{N}_c} \left\lVert\hat{\mathbf{x}}_c^t + \mathrm{R}_c^t\left(\bar{\mathbf{x}}_i^{1} - \hat{\mathbf{x}}_c^1 \right)-\bar{\mathbf{x}}_i^{t}\right\rVert^2 \,,
\end{equation}
$\mathcal{L}_\mathrm{ps}$ is a position smoothness loss
\begin{equation}
    \mathcal{L}_\mathrm{ps} = \sum_t \left\lVert \hat{\mathbf{x}}_c^{t-1} + \hat{\mathbf{x}}_c^{t+1} - 2 \hat{\mathbf{x}}_c^t \right\rVert^2 \,,
\end{equation}
and $\mathcal{L}_\mathrm{rs}$ is a rotation smoothness loss
\begin{equation}
    \mathcal{L}_\mathrm{rs} = \sum_t \left\lVert \hat{\mathbf{q}}_c^{t-1} + \hat{\mathbf{q}}_c^{t+1} - 2 \hat{\mathbf{q}}_c^t \right\rVert^2 \,.
\end{equation}
$\mathcal{N}_c$ the a set of indices of particles near the initial contact point location $\tilde{\mathbf{x}}_c^1$ and $\hat{\mathrm{R}}_c^t$ is a ratation matrix computed from $\hat{\mathbf{q}}_c^t$. In practice, we set $\lambda_\mathrm{ps}$ to 0.1 and $\lambda_\mathrm{rs}$ to $10^{-4}$.

M-PhyGs incorporates the estimated contact-point motion into the MPM simulation by adding boundary conditions. As directly imposing boundary conditions on particle positions is non-trivial, we instead impose boundary conditions on particle velocities $\mathbf{v}_{i,\mathrm{bc}}^t$
\begin{equation}
    \mathbf{v}_{i,\mathrm{bc}}^t = \frac{\mathbf{x}_{i,\mathrm{c}}^{t+1} - \mathbf{x}_{i,\mathrm{c}}^{t}}{n_s T} + \frac{\mathbf{x}_{i,\mathrm{c}}^t - \mathbf{x}_i^t}{\kappa T} \,,
\end{equation}
where $\mathbf{x}_{i,\mathrm{c}}^{t}$ is the position of the particle computed from the recovered contact point motion under the rigid-motion assumption, $T$ is the simulation substep size, $n_s$ is the number of substeps per frame (\ie, $n_s T$ is the time interval between neighboring frames), and $\kappa$ is a hyperparameter. The first term is the particle velocity computed from the estimated contact point motion, and the second term encourages the particle location to become close to $\mathbf{x}_{i,\mathrm{c}}^t$. For particles near the estimated contact point, at each simulation substep, we overwrite the particle velocity $\mathbf{v}_i^t$ with $\mathbf{v}_{i,\mathrm{bc}}^t$ before the particle-to-grid transfer in the MLS-MPM simulation~\cite{hu2018mlsmpm}.

\section{Implementation Details}
\paragraph{Initial Particle Registration}
For video sequences capturing the object dynamics, we align the physical particles and the 3D Gaussians to the first frames of the sequences based on sparse correspondences and a rendering loss. We first manually annotate 2D-2D correspondences between the first frames and frames corresponding to the rest shape, and then recover 3D-3D correspondences by triangulation. From these 3D-3D correspondences, we estimate a rigid motion that roughly aligns the physical particles and the 3D Gaussians with the first frames. We further refine per-particle 3D positions by minimizing a rendering loss that is a weighted sum of the RGB loss $\mathcal{L}_\mathrm{rgb}$ and the DINO feature loss $\mathcal{L}_\mathrm{DINO}$. During the refinement, we impose a local rigidity loss~\cite{luiten2023dynamic} with a large weight to preserve the overall structure of the particle distribution.

\paragraph{Temporal Mini Batching}
During the material parameter estimation, the temporal mini-batch size is gradually increased. We start with a mini-batch size of 5 frames and then increase it to 10, 15, and 25 frames. The number of mini-batches is set to 3 for batch sizes up to 15 and reduced to 2 when the batch size is 25. During this temporal mini-batch training, the 3D loss $\mathcal{L}_\mathrm{3D}$ is minimized. At the end, the material parameters are refined using all 50 frames without temporal mini-batching. During the refinement, the 3D loss and then the 2D loss $\mathcal{L}_\mathrm{2D}$ are minimized. For each stage, the number of iterations is 150.

\paragraph{MLS-MPM Simulation}
We implemented our differentiable MLS-MPM simulator based on the implementation originally implemented by Xie \etal~\cite{xie24physgaussian} and later modified by Zhang \etal~\cite{zhang24physdreamer} with adaptations for integration with M-PhyGs. First, as described in \cref{sec:contact_point_motion}, the motion of the contact point is incorporated into the MPM simulation. The efficiency of the MPM simulator is also improved by constructing voxel grids for the particle-to-grid and grid-to-particle transfers only within the bounding box of the simulation particles.
For further details of the MLS-MPM simulation, please refer to the original paper~\cite{hu2018mlsmpm} and the SIGGRAPH course notes~\cite{jiang16mpmcourse}.

\paragraph{Experimental Settings}
We set the grid spacing for the particle-to-grid and grid-to-particle transfers in the MLS-MPM simulation to 11 mm. The spacing between uniformly distributed simulation particles is 2 mm, and the number of simulation particles ranges from 37k to 180k per object.
Most experiments were run on NVIDIA A100 GPUs with 80 GB of memory. For objects with 135k particles or more, we used an NVIDIA H200 GPU with 141 GB of memory. The total training time per object ranges from 80 to 100 hours. Inference (dynamics prediction) takes on the order of a few seconds per frame.

\section{Modifications to Existing Methods}
As most existing methods do not incorporate human-object interaction into their simulators, we adapt them for experimental comparison.

For PhysDreamer~\cite{zhang24physdreamer}, OmniPhysGS~\cite{lin25omniphysgs}, and Pixie~\cite{le2025pixie}, similar to \cref{sec:contact_point_motion}, we incorporate the contact point motion into their MPM simulators. Note that this modification does not affect their training procedures as they do not require real videos as inputs.
For PhysDreamer~\cite{zhang24physdreamer} and OmniPhysGS~\cite{lin25omniphysgs}, we use an image from the dense view capture as the input image. 
For Pixie~\cite{le2025pixie}, we use the dense view capture to recover the input voxel representation. 

For Spring-Gaus~\cite{zhong24springgaus}, we incorporate the contact point motion into its simulator, by replacing the positions of simulation particles near the contact point with those computed from the contact point motion. This replacement is performed just before the velocity update of each simulation substep. As the registration method of Spring-Gaus struggles with the complex geometry of flowers in the Phlowers dataset, we instead use our recovered Gaussians and registration results as inputs to its material parameter estimation stage.

For gs-dynamics~\cite{zhang2024dynamics}, it is difficult to recover 3D Gaussians from initial frames of the videos of Phlowers dataset, as they suffer from occlusions by a person. We instead recover 3D Gaussians for gs-dynamics from the dense view capture and align them with the first frame, similar to M-PhyGs. For each flower, we use a single 50-frame sequence for training, as in the training of M-PhyGs.

For GIC~\cite{cai24gic}, it is difficult to apply the original implementation to Phlowers dataset as it does not incorporate the contact point motion and requires a large amount of GPU memory for training for large objects (large flowers). We instead implement GIC on top of our M-PhyGs framework. We implement the Chamfer distance loss and the mask L1 loss of GIC and optimize a single set of material parameters, rather than our multi-material representation.

\section{Additional Experimental Results}

\begin{figure*}
    \centering
    \includegraphics[width=\textwidth, keepaspectratio]{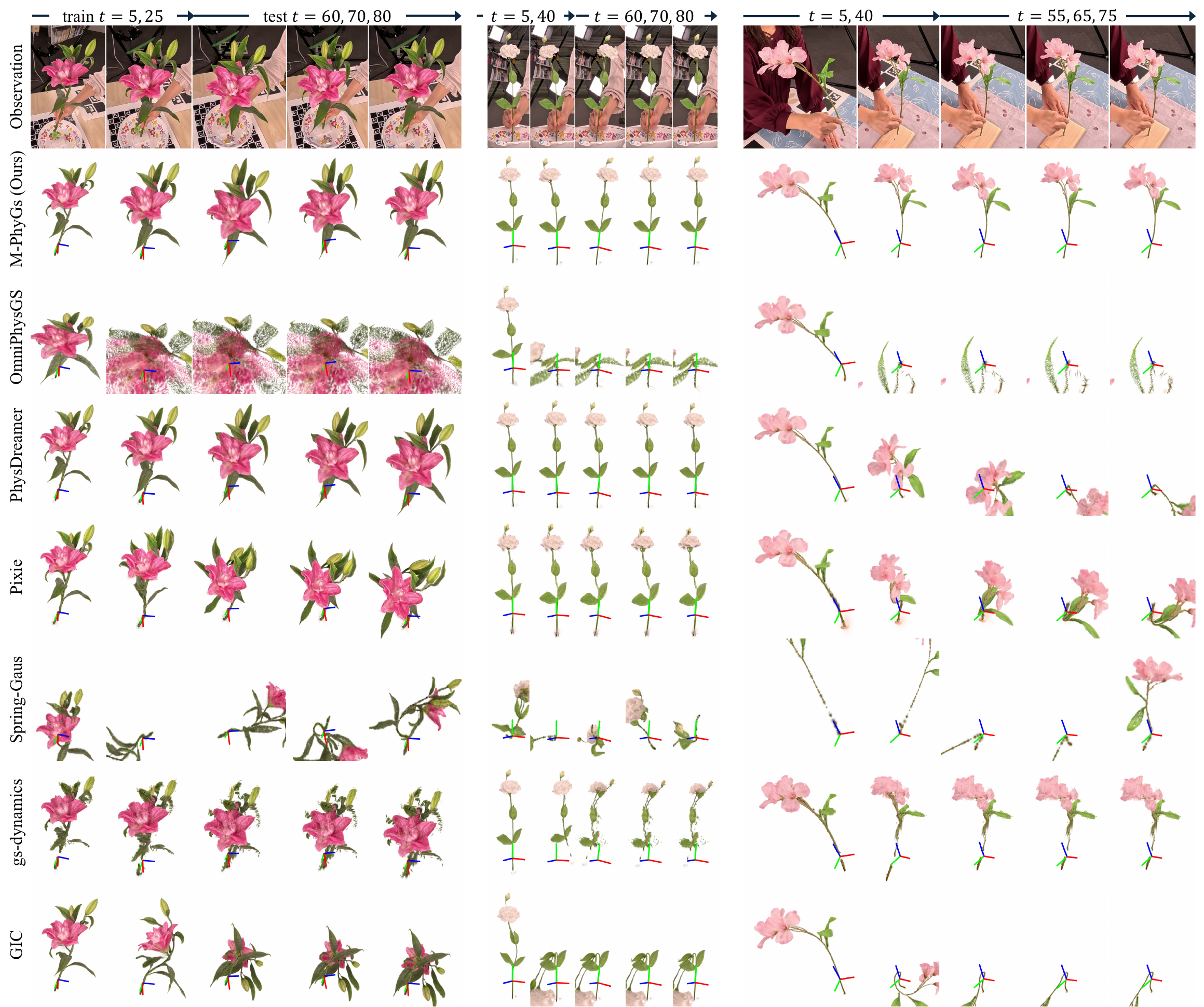}
    \caption{
    Unseen dynamics predicted with estimated physical material parameters. M-PhyGs successfully predicts the complex motion of each flower which aligns with the actual held out observations. 
    }
  \label{fig:result_future_prediction_supp_1}
\end{figure*}

\begin{figure*}
    \centering
    \includegraphics[width=\textwidth, keepaspectratio]{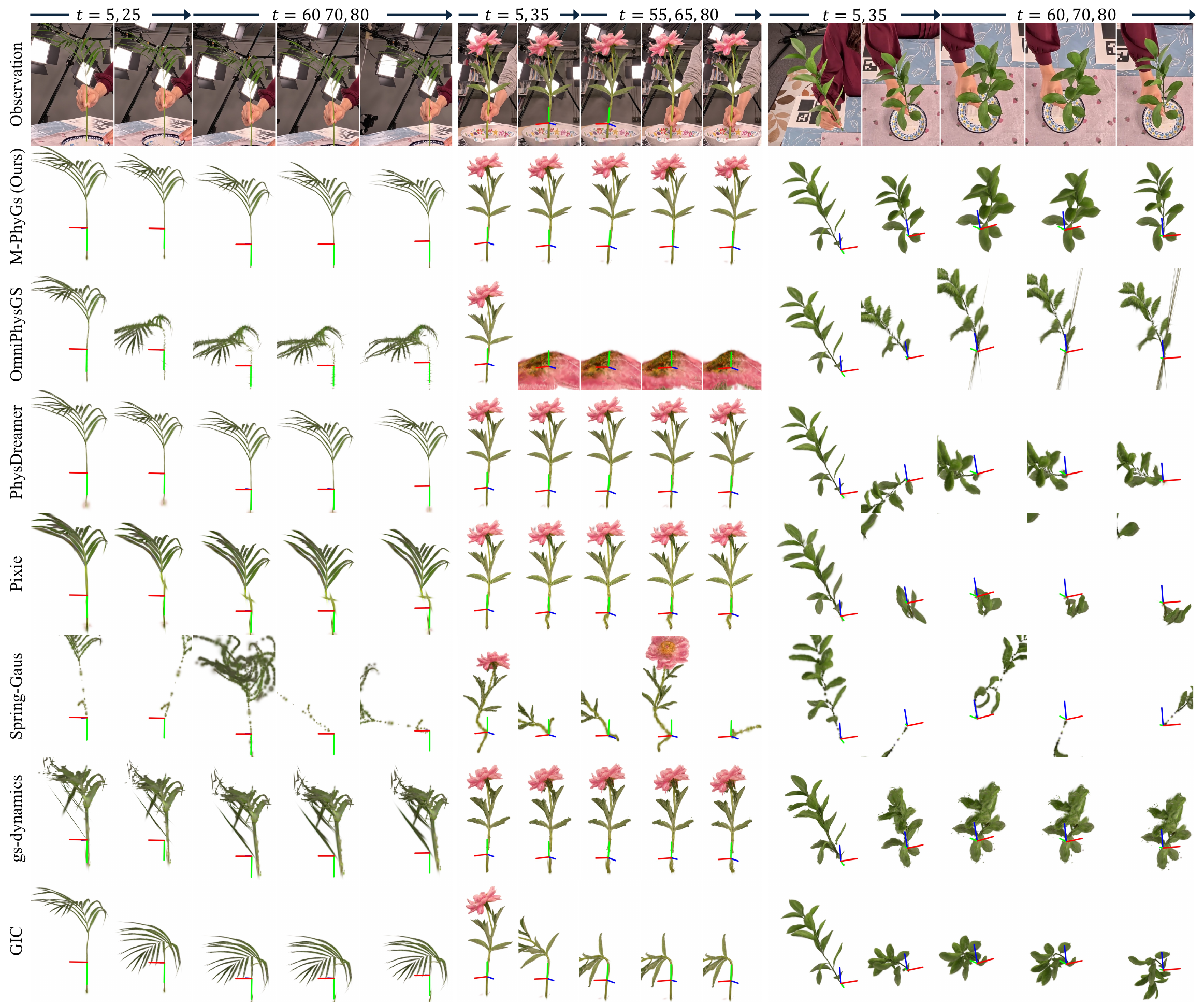}
    \caption{
    Unseen dynamics predicted with estimated physical material parameters. M-PhyGs successfully predicts the complex motion of each flower which aligns with the actual held out observations. 
    }
  \label{fig:result_future_prediction_supp_2}
\end{figure*}

\begin{figure*}

    \centering
    \includegraphics[width=\textwidth, keepaspectratio]{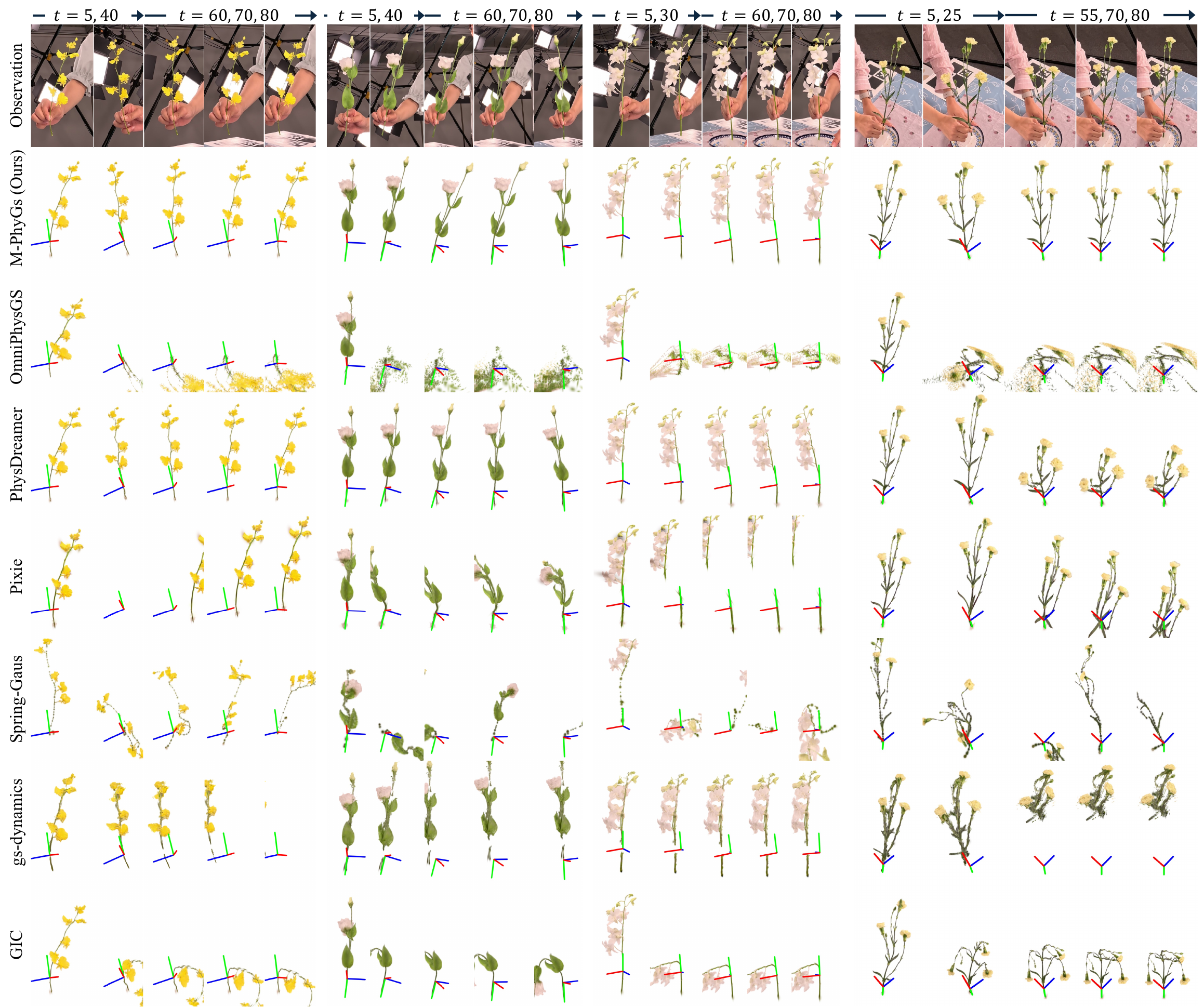}
    \caption{
    Unseen dynamics predicted with estimated physical material parameters. M-PhyGs successfully predicts the complex motion of each flower which aligns with the actual held out observations. 
    }
  \label{fig:result_future_prediction_supp_3}
\end{figure*}

\Cref{fig:result_future_prediction_supp_1,fig:result_future_prediction_supp_2,fig:result_future_prediction_supp_3} show additional qualitative results of dynamic prediction for unseen (in-sequence) interactions. 
The results show that M-PhyGs successfully predicts the complex motion of each flower which aligns with the actual held out observations. 

\begin{figure*}
    \centering
    \includegraphics[width=\textwidth, keepaspectratio]{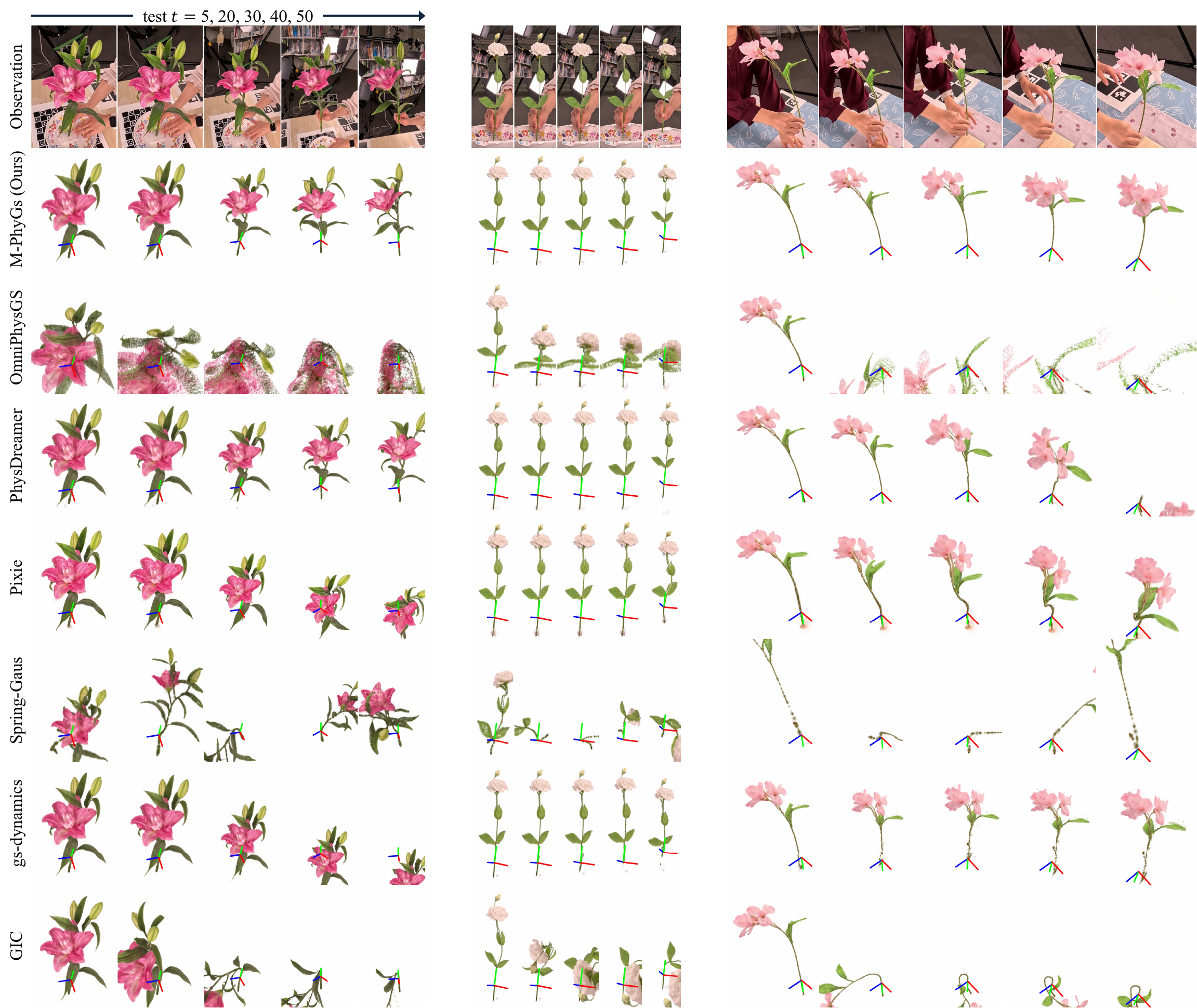}
    \caption{
    Qualitative results of cross-sequence dynamics prediction using estimated material. M-PhyGs successfully predicts the complex motion of each flower even in this cross-sequence setting. 
    }
  \label{fig:result_cross_sequence_prediction_1}
\end{figure*}

\begin{figure*}
    \centering
    \includegraphics[width=\textwidth, keepaspectratio]{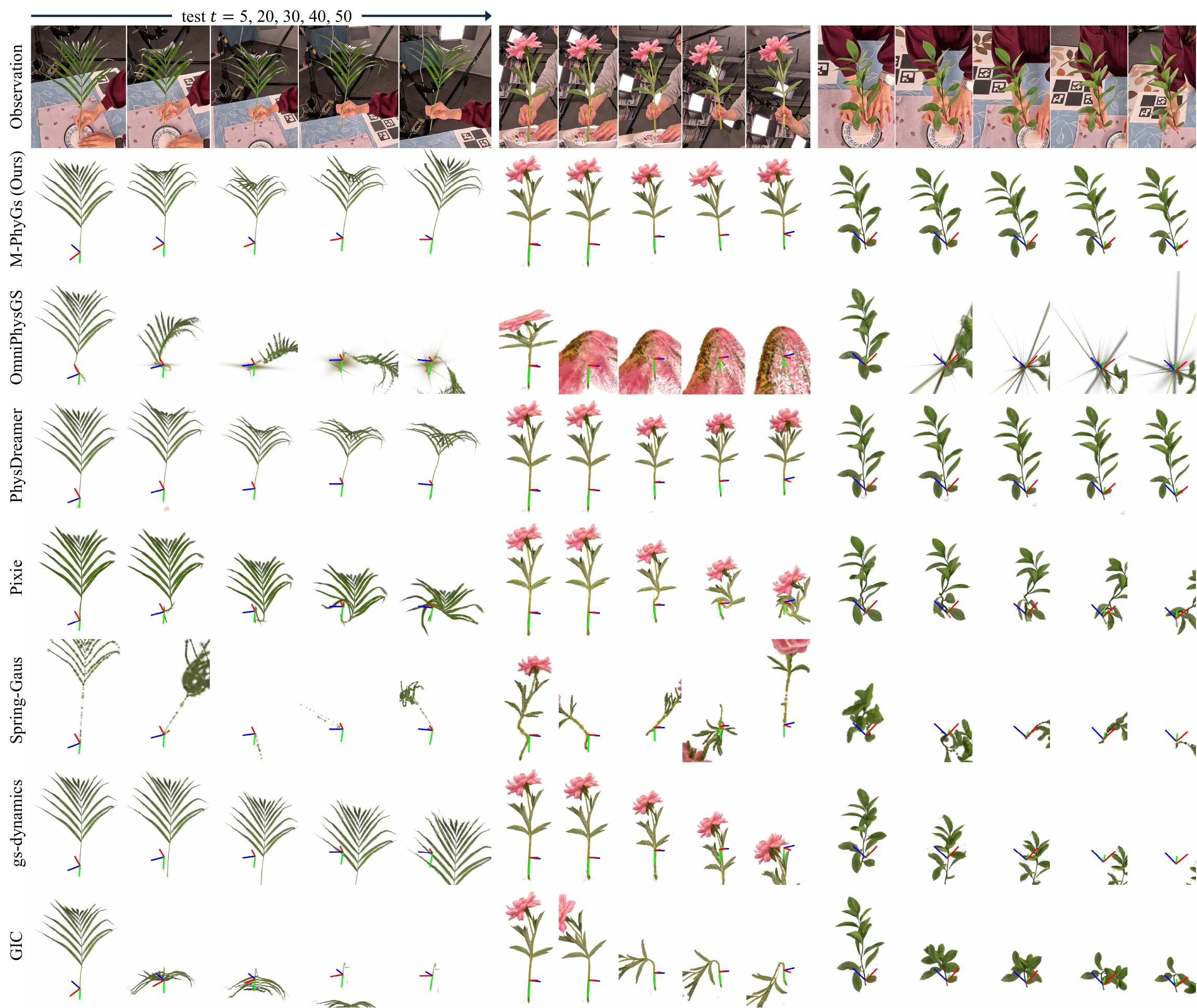}
    \caption{
    Qualitative results of cross-sequence dynamics prediction using estimated material. M-PhyGs successfully predicts the complex motion of each flower even in this cross-sequence setting. 
    }
  \label{fig:result_cross_sequence_prediction_2}
\end{figure*}

\begin{figure*}

    \centering
    \includegraphics[width=\textwidth, keepaspectratio]{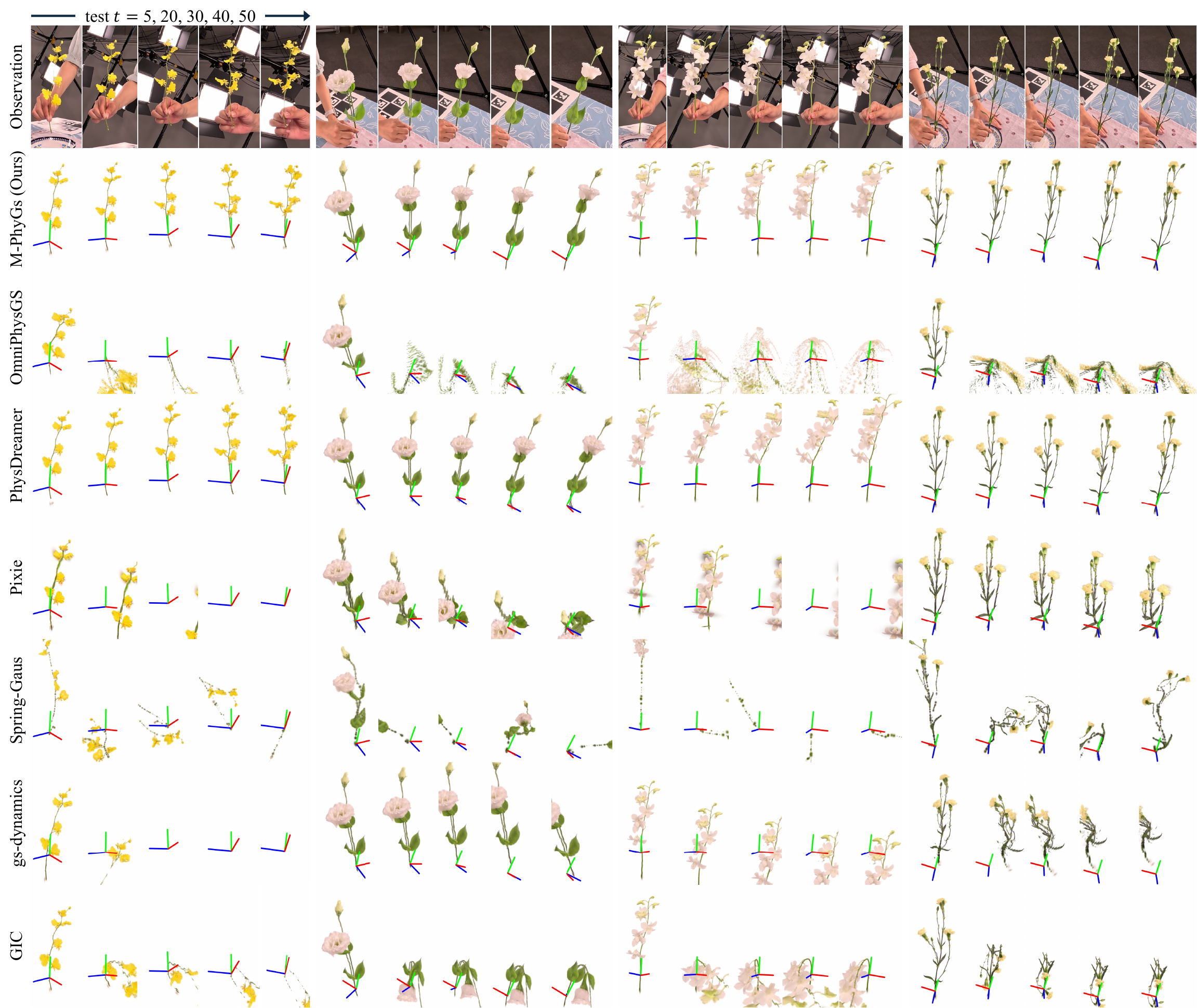}
    \caption{
    Qualitative results of cross-sequence dynamics prediction using estimated material. M-PhyGs successfully predicts the complex motion of each flower even in this cross-sequence setting. 
    }
  \label{fig:result_cross_sequence_prediction_3}
\end{figure*}

We also conduct cross-sequence dynamics prediction, \ie, estimate physical material parameters from one sequence of an object and use it to predict dynamics in another sequence of the same object. \Cref{fig:result_cross_sequence_prediction_1,fig:result_cross_sequence_prediction_2,fig:result_cross_sequence_prediction_3} show additional  qualitative results. The results show that M-PhyGs successfully predicts dynamics even in this cross-sequence setting, which further shows the accuracy of their physical material parameter estimates.

{
    \small
    \bibliographystyle{ieeenat_fullname}
    \bibliography{main}
}

\end{document}